\documentclass[11pt,a4paper]{article}
\usepackage[hyperref]{ranlp2021}
\usepackage{times}
\usepackage{latexsym}

\usepackage{booktabs}

\usepackage{microtype}
\usepackage{bm}
\usepackage{graphicx}
\usepackage{array}
\usepackage{multirow}

\aclfinalcopy 

\newcommand{\rashkincorpus}{\texttt{TSHP-17}}
\newcommand{\proppycorpus}{\texttt{QProp} }

\title{Interpretable Propaganda Detection in News Articles}

\author{Seunghak Yu$^{1}$\thanks{\quad Work conducted while the author was at MIT CSAIL.} \quad Giovanni Da San Martino$^{2}$ \\
        \textbf{Mitra Mohtarami}$^{3}$ \quad \textbf{James Glass}$^{3}$ \quad \textbf{Preslav Nakov}$^{4}$ \\
        $^{1}$ Amazon Alexa AI, Seattle, WA, USA \\
        $^{2}$ Department of Mathematics, University of Padova, Italy \\
        $^{3}$ MIT Computer Science and Artificial Intelligence Laboratory, Cambridge, MA, USA \\
        $^{4}$ Qatar Computing Research Institute, HBKU, Qatar \\
        \tt yuseungh@amazon.com, dasan@math.unipd.it \\
        {\tt \{mitra, glass\}@csail.mit.edu, pnakov@hbku.edu.qa}
    }

\date{}

\begin{document}
\maketitle
\begin{abstract}
Online users today are exposed to misleading and propagandistic news articles and media posts on a daily basis. To counter thus, a number of approaches have been designed aiming to achieve a healthier and safer online news and media consumption. Automatic systems are able to support humans in detecting such content; yet, a major impediment to their broad adoption is that besides being accurate, the decisions of such systems need also to be interpretable in order to be trusted and widely adopted by users. Since misleading and propagandistic content influences readers through the use of a number of deception techniques, we propose to detect and to show the use of such techniques as a way to offer interpretability. In particular, we define qualitatively descriptive features and we analyze their suitability for detecting deception techniques. We further show that our interpretable features can be easily combined with pre-trained language models, yielding state-of-the-art results.
\end{abstract}

\section{Introduction}

With the rise of the Internet and social media, there was also a rise of fake \cite{CIKM2020:FANG}, biased \cite{baly-etal-2020-detect,baly2020written}, hyperpartisan \cite{potthast2018stylometric}, and propagandistic content \cite{emnlp2019:fine:grained}. In 2016, news got weaponized, aiming to influence the US Presidential election and the Brexit referendum, making the general public concerned about the dangers of the proliferation of fake news \cite{howard2016bots,faris2017partisanship,lazer1094,Vosoughi1146,bovet2019influence}. 

There ware two reasons for this. First, disinformation disguised as news created the illusion that the information is reliable, and thus people tended to lower their barrier of doubt compared to when information came from other types of sources. 

Second, the rise of citizen journalism led to the proliferation of various online media, and the veracity of information became an issue. In practice, the effort required to fact-check the news, and its bias and propaganda remained the same or even got more complex, compared to traditional media, since the news was re-edited and passed through other media channels.

\begin{figure}
    \centering
	\includegraphics[width=\columnwidth]{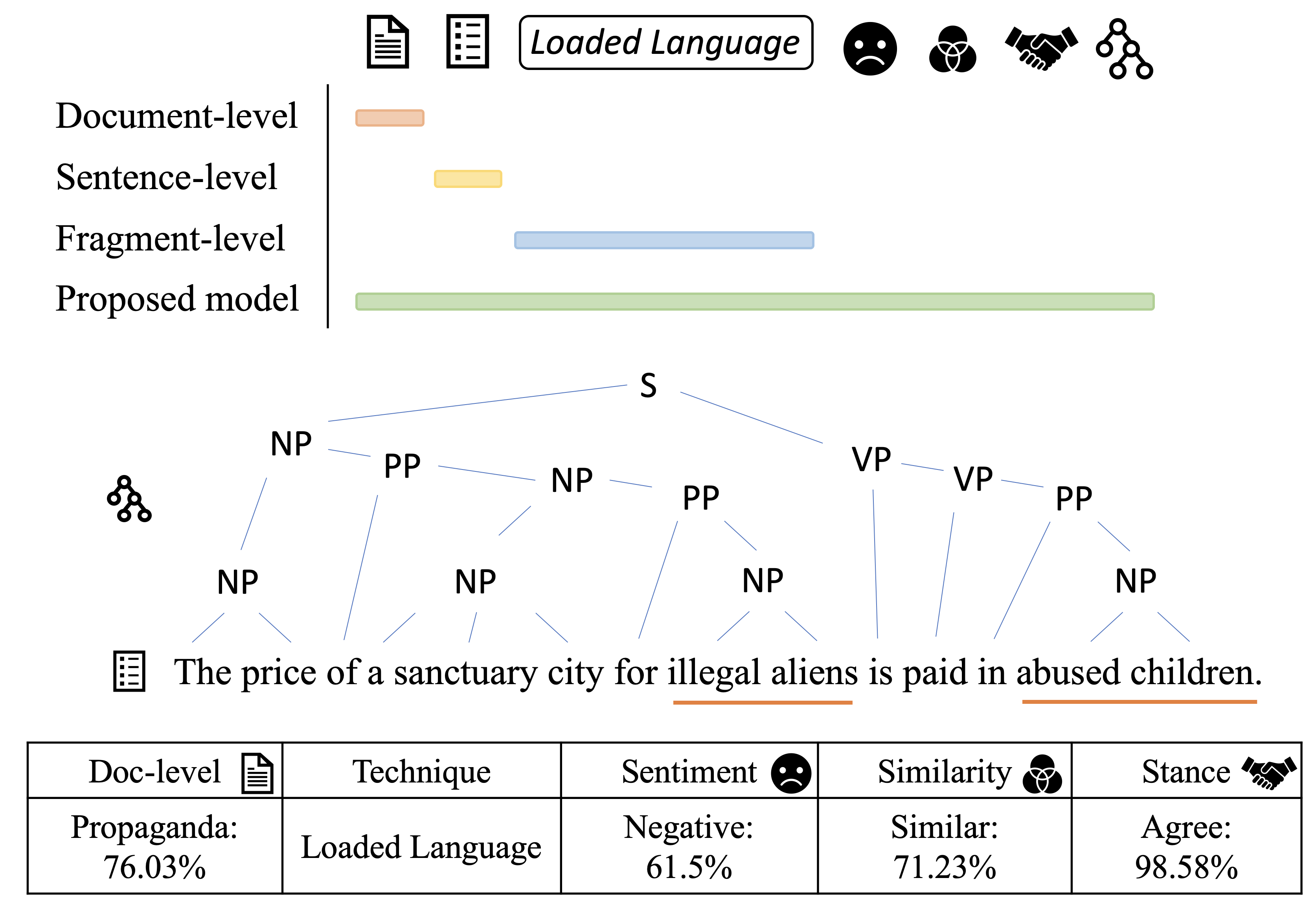}
    \caption{Comparison of propaganda prediction interpretability using existing methods. Our proposed method helps users to interpret propaganda predictions across various dimensions, e.g.,~is there a lot of positive/negative sentiment (can signal the use of \emph{loaded language}, which appeals to emotions), are the target sentence and the document body related to the title, does the sentence agree/disagree with the title, etc. Each symbol in the top bar chart represents an information source for propaganda detection. \label{fig:compare}}
\end{figure}

Propaganda aims to influence the audience with the aim of advancing a specific agenda~\cite{survey2020}. Detecting it is tricky and arguably more difficult than finding false information in an article. This is because propagandistic articles are not intended to simply make up a story with objective errors, but instead use a variety of techniques to convince people, such as selectively conveying facts or appealing to emotions~\cite{jowett2012propaganda}.

While many techniques are ethically questionable, we can think of propaganda techniques as rhetorical expressions that effectively convey the author's opinion~\cite{o2004politics}. Due to these characteristics, propagandistic articles are often produced primarily for political purposes (but are also very common in commercial advertisement), which directly affect our lives, and are commonly found even in major news media outlets, which are generally considered credible. 

The importance of detecting propaganda in the news has been recently emphasized, and research is being conducted from various perspectives~\cite{rashkin2017truth,barron2019proppy,emnlp2019:fine:grained}. However, while previous work has done reasonable job at detecting propaganda, it has largely ignored the question of why the content is propagandistic, i.e.,~there is a lack of interpretability of the system decisions, and in many cases, there is a lack of interpretability of the model as well, i.e.,~it is hard to understand what the model actually does even for its creators.

Interpretability is indispensable if propaganda detection systems are to be trusted and accepted by the users. According to the confirmation bias theory~\cite{nickerson1998confirmation}, people easily accept new information that is consistent with their beliefs, but are less likely to do so when it contradicts what they already know. Thus, even if a model can correctly predict which news is propagandistic, if it fails to explain the reason for that, people are more likely to reject the results and to stick to what they want to believe. In order to address this issue, we propose a new formulation of the propaganda detection task and a model that can explain the prediction results. Figure~\ref{fig:compare} compares the coverage of the explanations for pre-existing methods vs. our proposal. 

Our contributions can be summarized as follows:

 \begin{itemize}
    \item We study how a number of information sources relate to the presence and the absence of propaganda in a piece of text.
    \item  Based on this, we propose a general framework for interpretable propaganda detection.
    \item We demonstrate that our framework is complementary to and can be combined with large-scale pre-trained transformers, yielding sizable improvements over the state of the art.  
 \end{itemize}

\section{Task Setup}
\label{s:setup}

Given a document $\boldsymbol{d}$ that consists of $n$ sentences $\boldsymbol{d}=\{d_{i}\}^{n}_{i=1}$, each sentence should be classified as belonging to one of 18 propaganda techniques or as being non-propaganda. The exact definition of propaganda can be subtly different depending on the social environment and the individual's growth background, and thus it is not surprising that the propaganda techniques defined in the literature differ~\cite{miller1939techniques,jowett2012propaganda,hobbs2014teaching,torok2015symbiotic,weston2018rulebook}. The techniques we use in this paper are shown in Table~\ref{tab:prop_classes}. \citet{emnlp2019:fine:grained} derived the propaganda techniques from the literature: they selected 18 techniques and manually annotated 451 news articles with a total of 20,110 sentences. This dataset\footnote{\url{http://propaganda.math.unipd.it/}} has fragment-level labels that can span over multiple sentences and can overlap with other labeled spans.

This granular labeling went beyond our scope and we had to restructure the data. First, we divided the data into sentences. Second, in order to reduce the complexity of the task, we changed the multi-label setup to a multi-class one by ignoring duplicate labels and only allowing one technique per sentence (the first one), breaking ties at random. As a result, we obtained 20,111 sentences labeled with a non-propaganda class or with one of 18 propaganda techniques. Based on this data, we built a system for predicting the use of propaganda techniques at the sentence level, and we provided the semantic and the structural information related to propaganda techniques as the basis of the results.

\begin{table}
\centering
\footnotesize
\begin{tabular}{@{}p{2cm}p{5cm}@{}}  
\toprule
\bf Techniques  & \bf Definition \\
\midrule
Name Calling & give an object an insulting label \\
Repetition & inject the same message over and over\\
Slogans & use a brief and memorable phrase \\
Appeal to Fear& plant fear against other alternatives \\
Doubt & questioning the credibility \\
Exaggeration & exaggerate or minimize something \\
Flag-Waving & appeal to patriotism \\
LL & appeal to emotions or stereotypes \\
RtoH & the disgusted group likes the idea\\
Bandwagon & appeal to popularity \\
CO & assume a simple cause for the outcome \\
OIC & use obscure expressions to confuse \\
AA & use authority's support as evidence \\
B\&W Fallacy & present only two options among many\\
TC & discourage meaningful discussion \\
Red Herring & introduce irrelevant material to distract \\
Straw Men & refute a nonexistent argument \\
Whataboutism & discredit an opponent’s position\\
\bottomrule
\end{tabular}
\caption{List of propaganda techniques and brief definitions. LL: Loaded Language, RtoH: Reduction to Hitlerum, CO: Casual Oversimplification, OIC: Obfuscation, Intentional vagueness, Confusion, AA: Appeal to Authority, TC: Thought-terminating Clich\'es.}
\label{tab:prop_classes}
\end{table}

\section{Proposed Method\label{sec:features}}

Our method can detect the propaganda for each sentence in a document, and can explain what propaganda technique was used with interpretable semantic and syntactic features. We further propose novel features conceived in the study of human behavioral characteristics. More detail below.

\subsection{People Do Not Read Full Articles}

Behavior studies have shown that people read less than 50\% of the articles they find online, and often stop reading after the first few sentences, or even after the title \cite{manjoo2013you}. Indeed, we found that 77.5\% of our articles use propaganda techniques in the first five sentences, 65\% do so in the first three sentences, and 31.07\% do so in the title.

We used three types of features ($\boldsymbol{f}^{rp}$, ${\boldsymbol{f}}^{sim}$, $\boldsymbol{f}^{stn}$) to account for these observations, which we describe below.

\subsubsection{Relative Position of the Sentence}
We define the relative position of a sentence as $\boldsymbol{f}^{rp}_i = i/n$, where $i$ is the sequence number of the sentence, and $n$ is the total number of sentences in the article.

\subsubsection{Topic Similarity and Stance with Respect to the Title}  

The title of an article typically contains the topic and also the author's view of that topic. Thus, we hypothesize that propaganda should also focus on the topic expressed in the title. 

We represent the relationship between the target sentence and the title by measuring the semantic similarity ${\boldsymbol{f}}^{sim}_i$ between them as the cosine between the sentence-BERT representations ($\boldsymbol{\phi}(x)$)~\cite{reimers2019sentence} of the target sentence $d_i$ and of the title $d_1$.

\begin{equation}
\boldsymbol{f}^{sim}_i = \frac{\boldsymbol{\phi}(d_1)\cdot \boldsymbol{\phi}(d_i)}{|\boldsymbol{\phi}(d_1)||\boldsymbol{\phi}(d_i)|}
\end{equation}

We further model the stance of a target sentence with respect to the title $\boldsymbol{f}^{stn}_i$ using a distribution over five classes: \emph{related}, \emph{unrelated}, \emph{agree}, \emph{disagree}, and \emph{discuss}. For this, we use a BERT model \cite{fang2019neural} fine-tuned on the Fake News Challenge dataset \cite{hanselowski-etal-2018-retrospective}.

The class \textit{unrelated} indicates that the sentence is not related to the claim made in the title, while \textit{agree} and \textit{disagree} refer to the sentence agreeing/disagreeing with the title, and finally \textit {discuss} is assigned when the topic is the same as that in the title, but there is no stance. We further introduce the \textit{related} class as the union of \textit{agree}, \textit{disagree}, and \textit{discuss}. We use as features the binary classification labels and also the probabilities for these five classes.

\subsection{Syntactic and Semantic Information}

Some propaganda techniques have specific structural or semantic characteristics. For example, \textit{Loaded Language} can be configured to elicit an emotional response, usually using an emotional noun phrase. To model this, we define the following three features: $\boldsymbol{f}^{dp}$, $\boldsymbol{f}^{sent}$, and $\boldsymbol{f}^{doc}$.

\subsubsection{Syntactic Information}

\begin{table}
\centering
\begin{tabular}{ll}  
\toprule
\textbf{Level}  & \textbf{Phrases} \\
\midrule
Clause       & S, SBAR, SBARQ, SINV, SQ\\
\midrule
Phrase    & ADJP, ADVP, CONJP, FRAG, INTJ, \\
        & LST, NAC, NP, NX, PP, PRN, PRT, \\
        & QP, RRC, UCP, VP, WHADJP, \\
        & WHAVP, WHADVP, WHNP, WHPP, X \\
\bottomrule
\end{tabular}
\caption{The syntactic labels we used as features.}
\label{tab:dp_annotations}
\end{table}

We used a syntactic parser to extract structural features about the target sentence $\boldsymbol{f}^{dp}_i$. Our hypothesis is that such information could help to discover techniques that have specific structural characteristics such as \emph{Doubt} and \emph{Black and White Fallacy}. We considered a total of 27 clause-level classes and phrase-level labels, including the \emph{unknown} class. The set is shown in Table~\ref{tab:dp_annotations}. 

\subsubsection{Sentiment of the Sentence}

The sentiment of the sentence $\boldsymbol{f}^{sent}_i$ is another important feature for detecting propaganda. This is because many propagandistic articles try to convince the readers by appealing to their emotions and prejudices. Thus, we extract the sentiment using a sentiment analyzer trained on social media data~\cite{hutto2014vader}, which gives a probability distribution over the following three classes: \emph{positive}, \emph{neutral}, and \emph{negative}. It further outputs \emph{compound}, which is a one-dimensional normalized, weighted composite score. We use all four scores as features.

\subsubsection{Document-Level Prediction}

If the document is likely to be propagandistic, then each of its sentences is more likely to contain propaganda. To model this, we use as a feature $\boldsymbol{f}^{doc}$ the score of the document-level propaganda classifier Proppy~\cite{barron2019proppy}. Note that Proppy is trained on articles labeled using media-level labels, i.e.,~using distant supervision. Therefore, all articles from a propagandistic source are considered to be propagandistic.

\section{Experimental Results\label{sec:exps}}

In this section, we present our experimental setup for interpretable propaganda detection and the evaluation results from our experiments. Specifically, we perform three sets of experiments: (\textit{i})~in Section~\ref{sec:featureanalysis}, we quantitatively analyze the effectiveness of the features we proposed in Section~\ref{sec:features}; (\textit{ii})~in Sections~\ref{sec:comparison} and \ref{sec:ablation}, we compare our feature-based model to the state-of-the-art model described in \cite{emnlp2019:fine:grained} using the experimental setup from that paper; (\textit{iii})~in Section~\ref{sec:detectingtechniques}, we analyze the performance of our model with respect to each of the 18 propaganda techniques.

\subsection{Quantitative Analysis of the Proposed Features\label{sec:featureanalysis}}

\begin{figure*}[hbtp]
    \centering
	\includegraphics[width=6.5in]{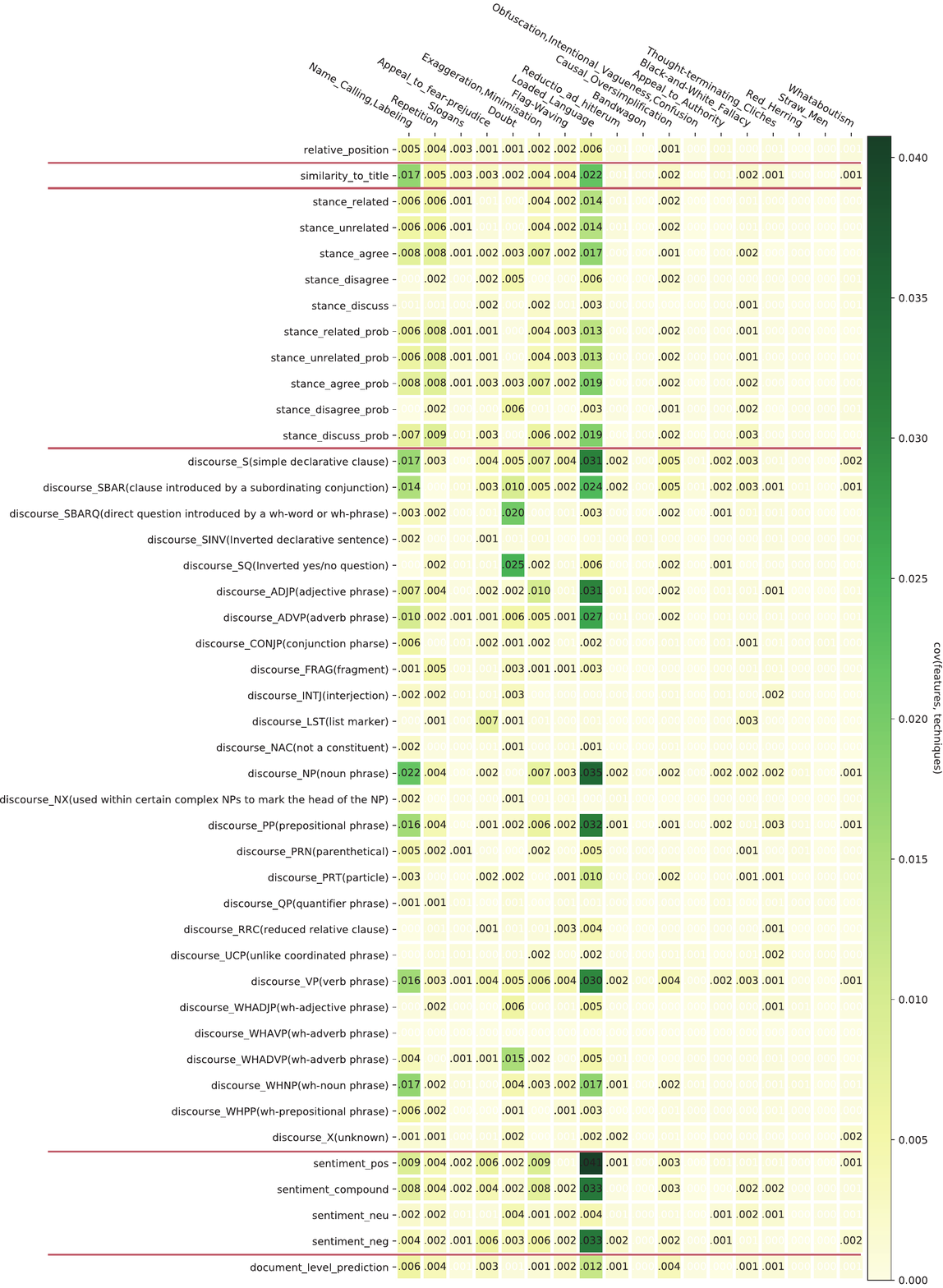}
    \caption{Covariance matrix between the 18 propaganda techniques and the proposed features. \label{fig:corr}}
\end{figure*}

Figure~\ref{fig:corr} shows the absolute value of the covariance between each of our features $\boldsymbol{f}$ and each of the 18 propaganda techniques $\boldsymbol{T}$. We calculated the values of the features on the training and on the development datasets, and we standardized their values. Then, we formulated this as a problem of calculating the covariance between continuous and Bernoulli random variables as follows: $cov(\boldsymbol{f},\boldsymbol{T}) = p\cdot(1-p)\cdot (E[\boldsymbol{f}|\boldsymbol{T}=1] - E[\boldsymbol{f}|\boldsymbol{T}=0])$.

The total number of sentences used is 16,137 (for the training and for the development datasets, combined), among which there are 4,584 propagandistic sentences. In Figure~\ref{fig:corr}, the vertical axis represents the proposed features, and the horizontal axis shows the individual propaganda techniques and the total number of instances thereof. Each square shows an absolute value of the covariance between some feature and some propaganda technique. We show absolute values in order to ignore the direction of the relationship, and we apply a threshold of 0.001 in order to remove the negligible relations from the figure.

Although the most frequent propaganda techniques appear in less than 10\% of the examples, they do show qualitatively meaningful associations. Indeed, we do not expect a feature to correlate with multiple techniques, as they are fundamentally different. We believe that having features that strongly correlate with one technique might be an advancement towards detecting that technique.

We can see that the structural information ($\boldsymbol{f}^{dp}$) and the sentiment of a sentence ($\boldsymbol{f}^{sent}$) are closely associated with certain propaganda techniques. For example, \emph{Loaded Language} has a strong correlation with features identifying words bearing either a positive or a negative sentiment. This makes sense as the authors are more likely to use emotional words rather than neutral ones, and \emph{Loaded Language} aims to elicit an emotional response. Similarly, \emph{Doubt} has high correlation with certain syntactic categories.

There are a number of interesting observations about the other features. For example, the relative position of sentences ($\boldsymbol{f}^{rp}$) is associated with more than half of the propaganda techniques. Moreover, the similarity to the title ($\boldsymbol{f}^{sim}$) and the stance with respect to the title ($\boldsymbol{f}^{stn}$) are strongly correlated with the likelihood that the target sentence is propagandistic. The features that indicate whether a sentence is related to the subject of the title are complementary, and thus the covariances are the same when absolute values are considered.

\subsection{Comparison to Existing Approaches \label{sec:comparison}}

Table~\ref{tab:performance} shows a performance comparison for our model vs. existing models on the sentence-level propaganda detection dataset~\cite{emnlp2019:fine:grained}. This dataset consists of 451 manually annotated articles, collected from various media sources, and a total of 20,111 sentences. Unlike the experimental setting in the previous sections, the task here is a binary classification one: given a sentence, the goal is to predict whether it contains \emph{at least one} of the 18 techniques or not. For the performance comparison, we used BERT~\cite{devlin2019bert}, which we fine-tuned for sentence-level classification using the Multi-Granularity Network (MGN)~\cite{emnlp2019:fine:grained} architecture on top of the [CLS] tokens (trained end-to-end), as this model improves the performance for both tasks by controlling the word-level prediction using information from the sentence-level prediction and vice versa.

We followed the original data split when training and testing the model, which is 14,137/2,006/3,967 for training/development/testing. We trained a Support Vector Machine (SVM) model\footnote{Ran on Intel Xeon E5-1620 CPU @ 3.60GHz x 4; 16GiB DDR3 RAM @ 1600MHz.} using the above-mentioned features and we optimized the values of the hyper-parameters on the development dataset using grid search. We used an RBF kernel with gamma=\{1e-3, \textbf{1e-4}\} and C=\{10,\textbf{100}\}.

We can see in Table~\ref{tab:performance} that our proposed model, which is based on interpretable features, performs relatively well when compared to fine-tuned BERT without direct semantic information about the target sentence. While our model is not state-of-the-art by itself, we managed to outperform the existing models and to improve over the state of the art by simply adding to it sentence embeddings as features~\cite{reimers2019sentence}, which were not fine-tuned on propaganda data. However, when the stance of the sentence and the embedding of the sentence are used together, performance decreases. This may be due to the two techniques based on semantic similarity being somewhat inconsistent.

\begin{table}
\begin{tabular}{p{0.48\columnwidth}*{3}{>{\centering\arraybackslash}p{0.1\columnwidth}}}
\toprule
\textbf{Model}  & \textbf{P} & \textbf{R} & \textbf{F1} \\
\midrule
fine-tuned BERT$^1$ & \textbf{63.20} & 53.16 & 57.74 \\
MGN$^1$ & 60.41 & 61.58 & 60.98 \\
\midrule
Proposed & 40.97 & 73.27 & 52.55 \\
Proposed w/ emb & 49.41 & 80.87 & 61.34 \\
Proposed w/ emb - $\boldsymbol{f}^{stn}$ & 49.59 & \textbf{81.44} & \textbf{61.64} \\
\bottomrule
\end{tabular}
\caption{Comparison of our method to pre-existing propaganda detection models at the sentence level for binary classification (\emph{propaganda} vs. \emph{non-propaganda}). The models flagged with $^1$ are described in~\cite{emnlp2019:fine:grained}. \label{tab:performance}}
\end{table}

\subsection{Ablation Study\label{sec:ablation}}

\begin{table}
\centering
\begin{tabular}{lccc}  
\toprule
\textbf{Ablations}  &  \textbf{Precision} & \textbf{Recall} & \textbf{F1} \\
\midrule
All & 40.97 & 73.27 & 52.55 \\
 - $\boldsymbol{f}^{rp}$ & 40.87 & 73.17 & 52.45 \\
- $\boldsymbol{f}^{sim}$ & 40.85 & 70.87 & 51.83 \\
- $\boldsymbol{f}^{stn}$ & 40.07 & 69.62 & 50.86 \\
- $\boldsymbol{f}^{dp}$ & 37.85 & 61.54 & 46.87 \\
- $\boldsymbol{f}^{sent}$ & 30.53 & 77.69 & 43.83 \\
\bottomrule
\end{tabular}
\caption{Ablation study for our model on binary propaganda detection at the sentence level.}
\label{tab:ablations}
\end{table}

Next, we performed an ablation study of the binary (propaganda vs. non-propaganda) model discussed in Section~\ref{sec:comparison}. 
The results are presented in Table~\ref{tab:ablations}. The values in the last row of the table, i.e., - $\boldsymbol{f}^{sent}$, are obtained by applying the document-level classifier, i.e.,~the feature $\boldsymbol{f}^{doc}$, to all sentences.
We can see that the structural information about the sentence ($\boldsymbol{f}^{dp}$) is the best feature for this task. This is due to the nature of some propaganda techniques that must have a specific sentence structure, such as \textit{Doubt}. In addition, as described above, since there are many techniques related to inducing emotional responses in the readers, it can be understood that the sentiment of a sentence may be a good feature, e.g.,~for \textit{Loaded Language}. These results are consistent with our findings in Section 4.1 above.
Moreover, the novel features we devised based on a human behavioral study for propaganda detection ($\boldsymbol{f}^{rp}$, $\boldsymbol{f}^{sim}$, $\boldsymbol{f}^{stn}$) improved the performance further. 
Overall, we can see in the table that all features contributed to the performance improvement. 

\subsection{Detecting the 18 Propaganda Techniques\label{sec:detectingtechniques}}

\begin{table}
\begin{tabular}{@{}lrrrr@{}}
\toprule
\textbf{Techniques}  & \textbf{P} & \textbf{R} & \textbf{F1} & \#\\
\midrule
Non-propaganda & 94.37 & 36.62 & 52.77 & 2,927\\
Name Calling & 14.16 & 21.92 & 17.20 & 146\\
Repetition & 4.60 & 5.59 & 5.05 & 143\\
Slogans & 3.75 & 20.69 & 6.35 & 29\\
Appeal to F.& 12.99 & 38.37 & 19.41 & 86\\
Doubt & 5.97 & 34.85 & 10.20 & 66\\
Exaggeration & 6.06 & 20.90 & 9.40 & 67\\
Flag-Waving & 10.98 & 44.62 & 17.63 & 65\\
Loaded L. & 32.80 & 20.13 & 24.95 & 303\\
Reduction& 8.00 & 22.22 & 11.76 & 9\\
Bandwagon & 0.00 & 0.00 & 0.00 & 3\\
Casual O. & 4.03 & 27.27 & 7.02 & 22\\
O, I, C & 0.00 & 0.00 & 0.00 & 5\\
Appeal to A. & 1.32 & 13.04 & 2.39 & 23\\
B\&W fallacy & 0.89 & 4.55 & 1.49 & 22\\
T. clich\'es & 3.67& 44.44 & 6.78 & 18\\
Red Herring & 0.00 & 0.00 & 0.00 & 11\\
Straw Men & 0.00 & 0.00 & 0.00 & 1\\
Whataboutism & 2.54 & 14.29 & 4.32 & 21\\
\bottomrule
\textbf{weighted avg} & 73.59 & 32.80 & \textbf{42.88} & 3,967\\
\hline
\end{tabular}
\caption{Performance of our proposed method for the task of detecting the 18 propaganda techniques, as evaluated at the sentence level.}
\label{tab:multi-class}
\end{table}

For the experiments described in the following, we revert back to the task formulation in Section~\ref{s:setup}, but we perform a more detailed analysis of the outcome of the model: for a given article, the system must predict whether each sentence uses propaganda techniques, and if so, which of the 18 techniques in Table~\ref{tab:prop_classes} it uses. 

Table~\ref{tab:multi-class} shows the performance of our model on this task. We can see in the rightmost column that some techniques appear only in a very limited number of examples, which explains the very low results for them, e.g.,~for \textit{Red Herring} and \textit{Straw Man}. In an attempt to counterbalance the lack of gold labels for some of the techniques, we used sentence embeddings with the proposed features to capture more semantic information. Since this task is more challenging than the binary classification problem, we can intuitively expect a performance reduction, resulting in a weighted average F1 score of 42.88. However, this formulation of the problem has the advantage of providing more granular predictions, thus enriching the propaganda detection results.

\section{Related Work\label{sec:relatedwork}}

Research on propaganda detection has focused on analyzing textual content \cite{BARRONCEDENO20191849,rashkin2017truth,emnlp2019:fine:grained,NLP4IF2019:propaganda:task,NeurIPS2019:propaganda,survey2020}. \citet{rashkin2017truth} developed the \rashkincorpus~corpus, which uses document-level annotation with four classes: \emph{trusted}, \emph{satire}, \emph{hoax}, and \emph{propaganda}. They trained a model using word $n$-gram representation and reported that the model performed well only on articles from sources that the system was trained on. \citet{BARRONCEDENO20191849} developed the \proppycorpus corpus with two labels: \textit{propaganda} vs. \textit{non-propaganda}. They also experimented on \rashkincorpus~and \proppycorpus~corpora, where for the \rashkincorpus~ corpus, they binarized the labels: \textit{propaganda} \textit{vs.} any of the other three categories. Similarly, \citet{Habernal.et.al.2017.EMNLP,Habernal2018b} developed a corpus with 1.3k arguments annotated with five fallacies, including \textit{ad hominem}, \textit{red herring}, and \textit{irrelevant authority}, which directly relate to propaganda techniques. Moreover, \citet{saleh-etal-2019-team} studied the connection between hyperpartisanship and propaganda.

A more fine-grained propaganda analysis was proposed by \citet{emnlp2019:fine:grained}, who developed a corpus of news articles annotated with 18 propaganda techniques which was used in two shared tasks: at SemEval-2020 \cite{DaSanMartinoSemeval20task11} and at NLP4IF-2020 \cite{NLP4IF2019:propaganda:task}. Subsequently, the Prta system was released \cite{da2020prta}, and improved models were proposed, addressing the limitations of transformers \cite{ECML2021:end:of:history}. The Prta system was used to perform a study of COVID-19 disinformation and associated propaganda techniques in Bulgaria \cite{RANLP2021:COVID19:Bulgarian} and Qatar \cite{RANLP2021:COVID19:Qatar}. Finally, multimodal content was explored in memes using 22 fine-grained propaganda techniques \cite{dimitrov2021detecting}, which was also used in a SemEval-2021 shared task \cite{SemEval2021-6-Dimitrov}.

\section{Conclusion and Future Work\label{sec:conclusions}}

We proposed a model for interpretable propaganda detection, which can explain which sentence in an input news article is propagandistic by pointing out the propaganda techniques used, and why the model has predicted it to be propagandistic. To this end, we devised novel features motivated by human behavior studies, quantitatively deduced the relationship between semantic or syntactic features and propaganda techniques, and selected the features that were important for detecting propaganda techniques. Finally, we showed that our proposed method can be combined with a pre-trained language model to yield new state-of-the-art results.

In future work, we plan to expand the dataset by creating a platform to guide annotators. The dataset will be updated continuously and released for research purposes.\footnote{\url{http://propaganda.qcri.org/}} We also plan to release an interpretable online system, with the aim to foster a healthier and safer online news environment.

\section*{Acknowledgements}

This research is part of the Tanbih mega-project,\footnote{\url{http://tanbih.qcri.org/}} which aims to limit the impact of ``fake news'', propaganda, and media bias by making users aware of what they are reading, thus promoting media literacy and critical thinking. It is developed in collaboration between the Qatar Computing Research Institute, HBKU and the MIT Computer Science and Artificial Intelligence Laboratory.

\bibliographystyle{acl_natbib}
\bibliography{ranlp2021}

\end{document}